\documentclass{article} 
\usepackage[]{acl2023}
\usepackage{times}

\usepackage{hyperref}
\usepackage{url}

\usepackage{subfigure}
\usepackage{graphicx}
\usepackage{amsthm}
\usepackage{amsmath}
\usepackage{bm}
\usepackage{bbding}

\usepackage{amsfonts,amssymb}
\usepackage{float}
\usepackage{algorithm}
\usepackage{algorithmic}
\usepackage[normalem]{ulem}

\usepackage{colortbl}
\usepackage{xcolor}

\usepackage{multicol}
\usepackage{multirow}

\title{Privacy-preserved LLM Cascade via CoT-enhanced Policy Learning}

\author{Kai Zhang$^{1}$\thanks{This research was conducted during the author's tenure as a student researcher at Google AI Innovation \& Research (AIR).}, Congchao Wang$^{2}$, Liqian Peng$^{2}$, Alec Go$^{2}$, Xiaozhong Liu$^{1}$ \\
$^{1}$Worcester Polytechnic Institute, $^{2}$Google AIR\\
\texttt{\{kzhang8, xliu14\}@wpi.edu} \\
\texttt{\{liqianp, congchaowang, ago\}@google.com} \\
}

\begin{document}

\maketitle


\begin{abstract}
    Large Language Models (LLMs) have gained significant attention in on-device applications due to their remarkable performance across real-world tasks. However, on-device LLMs often suffer from suboptimal performance due to hardware limitations. A promising solution to this challenge is cascading a weaker local (on-device) LLM with a more powerful server LLM. While existing research on LLM cascade primarily optimizes the performance-cost trade-off, real-world applications impose additional requirements, such as privacy preservation, which remain largely unaddressed. In this work, we move beyond existing confidence- and logit-based LLM cascade methods and propose $\mathbf{P^{3}Defer}$, a novel Chain-of-Thought (CoT)-enhanced \textbf{p}olicy learning framework for \textbf{p}rivacy-\textbf{p}reserved \textbf{defer}ral decision-making. Our approach effectively improves cascade efficiency while mitigating privacy risks. Extensive experiments on three benchmark datasets demonstrate the effectiveness and superiority of $\mathbf{P^{3}Defer}$ over existing methods.
\end{abstract}

\section{Introduction}
As Large Language Models (LLMs) continue to evolve rapidly \citep{touvron2023llama, achiam2023gpt, reid2024gemini}, they are increasingly being integrated into real-world applications, enhancing the intelligence of a wide range of systems. At the same time, mobile devices have become indispensable in everyday life. The emergence of on-device intelligence—such as Apple Intelligence \citep{gunter2024apple} and Gemini Live \citep{reid2024gemini}—which embeds LLMs directly into devices for more personalized and intelligent user interactions, is gaining traction but remains relatively underexplored \citep{xu2024device}. A major challenge in this area is the hardware limitations of mobile devices, including constraints on compute power, battery life, and storage capacity. As a result, only smaller LLMs, such as Gemma-2B \citep{team2024gemma}, can be deployed on these devices, leading to trade-offs in performance compared to larger, more powerful models like Gemini. This raises a critical question for the research community: how can we optimize on-device intelligence given these size constraints? The LLM cascade system presents a solution for this challenge.\par
\begin{figure*}[h]
    \centering
    \includegraphics[width=\textwidth]{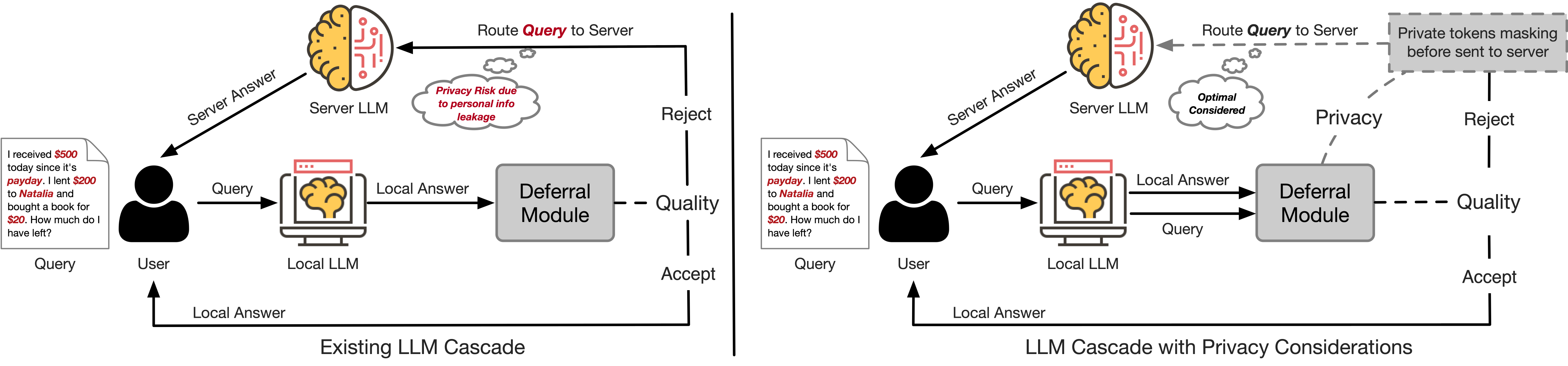}
    \caption{On the left is the existing LLM cascade, where the deferral module makes decisions solely based on the quality of the local answer, potentially leading to privacy leakage. On the right is the privacy-preserved LLM cascade, where deferral decisions are more aligned with the needs of real-world applications.}
    \label{fig:toy_example}
\end{figure*}
In an LLM cascade system, a query is usually first processed by a smaller, weaker local (on-device) LLM and is only escalated to a larger, stronger server LLM if the local model’s output is deemed insufficient by a deferral module, as shown in Figure \ref{fig:toy_example}. This paradigm has garnered significant attention recently \citep{chen2023frugalgpt, gupta2024language, yue2023large, wang2024cascade}. As larger LLMs are often substantially more expensive than their smaller counterparts (e.g., Gemini-1.5 Pro \citep{reid2024gemini} costs up to 35 times more than Gemini-Flash\footnote{https://ai.google.dev/pricing}), most existing LLM cascade works focused on the exploration of optimal trade-offs between cost and performance. However, real-world applications can be more complicated and requires the cascade system to make deferral decisions beyond just performance-cost consideration. For instance, privacy concerns may arise if personal data is routed to the server LLM where decisions are made based solely on the local answer's quality, as illustrated in Figure \ref{fig:toy_example}. Unfortunately, rare studies have explored the privacy-preserved LLM cascade system where to the best of our knowledge, only \citet{hartmann2024can} makes an attempt in this regards. In this study, we move beyond existing cascade system to make a pioneer step for including privacy concerns into the deferral decision making. 

One key focus of LLM cascade research is the design of deferral criteria, which determine whether a query needs to be routed to the server model. Existing study on this can be divied into two paradigms: confidence-based and logit-based methods (Please refer to Appendix \ref{sec:appendix_methodology} for more details if readers are not familiar with deferral decision making in LLM cascade.). Ideally, the deferral criteria should identify queries that the local LLM is unlikely to handle effectively, sending them to the server to significantly improve performance while keeping costs manageable. Conversely, sending queries that the local LLM can address with high quality to the server can result in unnecessary costs. Intuitively, model confidence could serve as a good indicator, with queries routed to the server when the local model is not confident with its response. For instance, \citet{zhu-etal-2024-towards} explored a self-critique strategy to leverage the local model's intelligence to produce a confidence level in terms of the local answer and make decisions based on the confidence level. However, \citet{jitkrittum2024does} noticed the weakness of confidence-based deferral rule in cases where distribution shifts occur between the training and test datasets. Logit-based methods step further by using the generated token logits of the local answer as features to make deferral decisions. For example, \citet{gupta2024language} found the length bias and token uncertainty problems in cascading by relying on the mean logits and proposed to leverage quantile logits as features to mitigate this problem. Additionally, \citet{wang2024cascade} introduced cascade-aware training, which incorporates both the local and server LLM’s logits into the loss function during local model training, helping the local LLM become more aware of which queries should be deferred to the server. Unfortunately, none of these works explored deferral decision making with respects to privacy concerns which aligns more with real-world needs. Moreover, both confidence-based and logit-based methods are by nature not feasible for including privacy considerations since logits and confidence can only reflect generation quality. To address this gap, we propose incorporating a policy learning strategy into the LLM cascade system. Instead of using a threshold for deferral decision making, we propose to train an agent that can make actions based on cascade needs. Moreover, Chain-of-Though (CoT) has been proven efficient in both training and training-free methods\citep{wu2024rethinking, yan2023ask}. Taking advantages, we propose a novel CoT-enhanced \textbf{p}olicy learning framework coupled with a private memory for better \textbf{p}rivacy-\textbf{p}reserved \textbf{defer}ral decision making ($\mathbf{P^{3}Defer}$). Different from logit-based or confidence-based methods, our $\mathbf{P^{3}Defer}$ leverages an agent to make deferral decisions (actions) based on the cascde system needs (environment). This paradigm enable our method to improve the LLM cascade performance while mitigate the privacy leakage problem.
In tandem, the contributions of this study are three-fold:\par
$\bullet$ We extend the current focus of LLM cascading beyond the traditional cost-performance trade-off to include privacy considerations, better aligning with the needs of real-world applications. \par
$\bullet$ We reformulate the LLM cascade task and innovatively incorporate a CoT-enhanced policy learning strategy coupled with a private objective to perform privacy-preserved deferral decision making, which provides a fresh perspective to the community. \par  
$\bullet$ Extensive experiments on three benchmarks have validated the efficiency and superiority of proposed $\mathbf{P^{3}Defer}$, witnessing improvements in LLM cascade performance while mitigating the privacy leakage\footnote{To encourage further explorations by the community, we will open-source our implementations (a copy is attached with this submissions).}. \par

\section{Methodology}
\begin{figure}[h]
    \centering
    \includegraphics[width=0.5\textwidth]{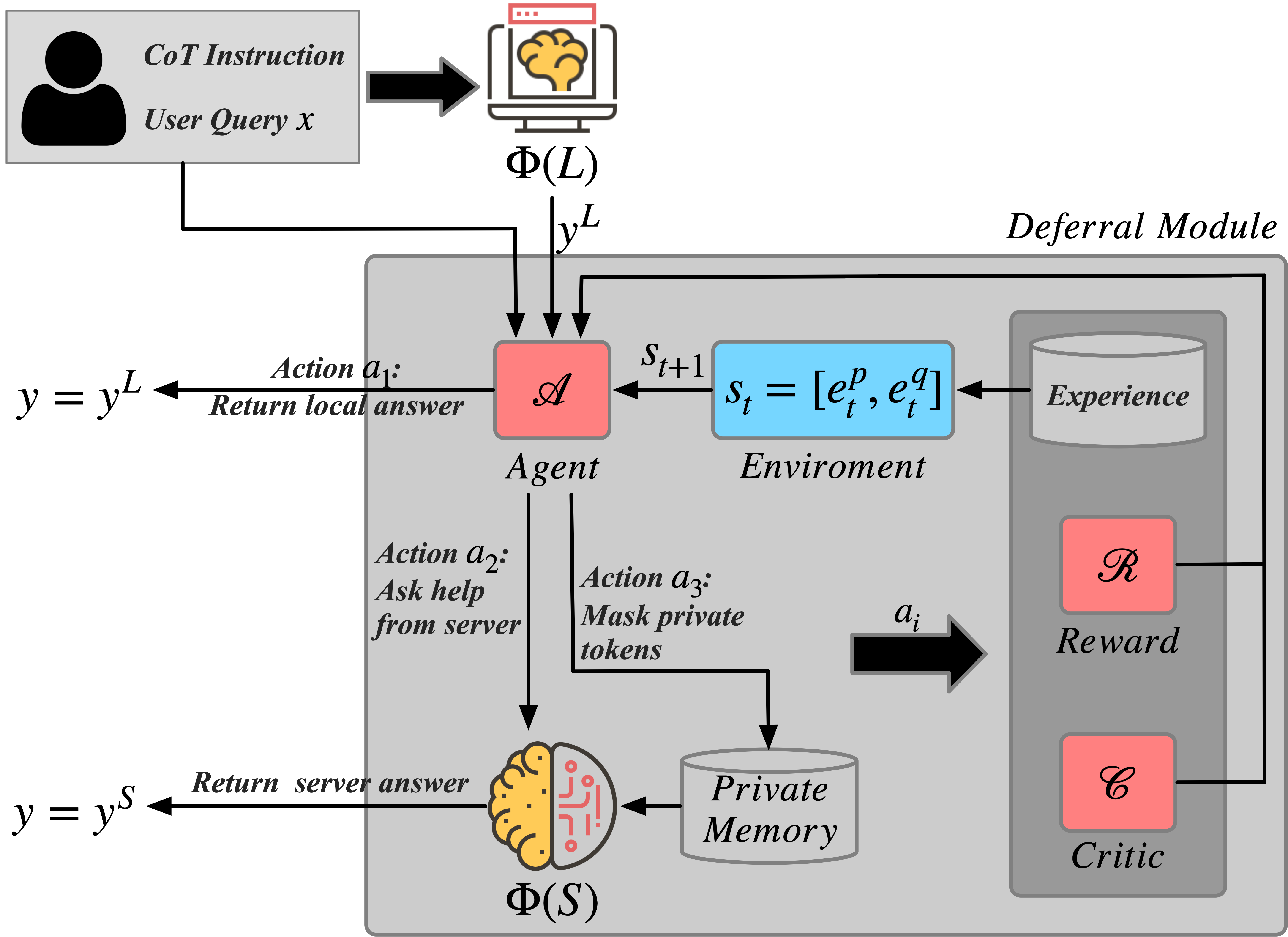}
    \caption{Overview of the proposed \(\mathbf{P^{3}Defer}\) framework. Given a user query \(x\), the local model \(\Phi(L)\) generates a response \(y^L\). The agent \(\mathcal{A}\) decides among three actions based on the state \(s_t\): (1) return \(y^L\), (2) defer to the server model \(\Phi(S)\) for response \(y^S\), or (3) mask private tokens via private memory. The agent is trained via reinforcement learning, where the reward function \(\mathcal{R}\) evaluates response quality and privacy, and the critic function \(\mathcal{C}\) assesses long-term decision-making.}
    \label{fig:methodology}
\end{figure}

\begin{algorithm}[ht]
\caption{$\mathbf{P^{3}Defer}$}
\label{alg:p3defer}
\begin{algorithmic}[1]
\REQUIRE $\Phi(L)$, $\Phi(S)$, $D(\mathcal{A}, \mathcal{R}, \mathcal{C}, \mathcal{S})$, $\mathcal{D}$, $\mathcal{M}$
\STATE Initialize policy network $\pi_{\theta}$, value function $V_{\psi}$, and experience buffer $\mathcal{B}$
\FOR{each training iteration}
    \FOR{each query $x \sim \mathcal{D}$}
        \STATE Generate local prediction $y^L = \Phi(L)(x)$
        \STATE Encode state $s_t = [e_t^p, e_t^q]$
        \STATE Sample action $a_t \sim \pi_{\theta}(s_t, x, y^{L})$
        \IF{$a_t = a_1$}
            \STATE Set final response $y = y^L$
            \STATE Compute reward $r_{t}$ using formula \ref{eq:reward_function}
        \ELSIF{$a_t = a_2$}
            \STATE Query server LLM: $y^S = \Phi(S)(x)$
            \STATE Set final response $y = y^S$
            \STATE Compute reward $r_{t}$ using formula \ref{eq:reward_function}
        \ELSIF{$a_t = a_3$}
            \STATE Mask sensitive tokens in $x$ via $\mathcal{M}$
            \STATE Generate modified query $x'$
            \STATE Query server LLM: $y^S = \Phi(S)(x')$
            \STATE Set final response $y = y^S$
            \STATE Compute reward $r_{t}$ using formula \ref{eq:reward_function}
        \ENDIF
        \STATE Store $(s_t, a_t, r_{t})$ in buffer $\mathcal{B}$
    \ENDFOR
    \STATE \textbf{Policy Update:}
    \STATE Compute advantage $\hat{A}_t =Q^\pi(s_t, a_t) - V^{\pi}(s_t)$
    \STATE Update policy $\theta \leftarrow \theta + \eta \nabla_{\theta} \mathbb{E} [\hat{A}_t \log \pi_{\theta}(a_t | s_t)]$
\ENDFOR
\end{algorithmic}
\end{algorithm}

\subsection{Preliminary Formulation}
Before proceeding, we will first present the preliminary concepts and formulations. Assuming we have an LLM cascade system consists of a local on-device LLM $\Phi(L)$ (smaller and weaker), a server LLM $\Phi(S)$ (larger and stronger) and a deferral module $D(\cdot)$. When a user sends a query $x$ to $\Phi(L)$ and the local model generates an initial answer $y^{L}$, the deferral module $D(\cdot)$ needs to determines whether it is necessary to invoke $\Phi(S)$ for the final response back to user. \\
Typically, existing methods use either the logit distribution of $y^{L}$ or prompting $\Phi(L)$ to do the deferral decision making\citep{wang2024cascade, zhu-etal-2024-towards}. That is say if $D(\cdot)$ accepts $y^{L}$, it becomes the final answer $y$ returned to the user. If rejected, the query $x$ is routed to $\Phi(S)$, and the server-generated answer $y^{S}$ serves as the final response $y$. However, these attempts are limited in incorporating real-world requirements into considerations due to the nature that their deferral modules are fixed and only make decisions based on confidence or logits\footnote{Please refer to appendix \ref{sec:appendix_preliminary} and \ref{sec:appendix_methodology} to check detailed explanations and preliminary results of existing methods.}. To step further, we move beyond and reformulate the deferral module into a trainable agent so that more considerations such as privacy can be added into deferral decision making.\\
In details, we can represent the deferral module by a tuple $D(\mathcal{A}, \mathcal{R}, \mathcal{C}, \mathcal{S}, x, y^{L})$ where $\mathcal{A}$ is the action space containing the actions that the deferral agent can take; $\mathcal{R}$ and $\mathcal{C}$ are the reward function and critic function, respectively; $\mathcal{S}$ denotes the set of environment states. A policy network $\pi_{\theta}: (\mathcal{S}, x, y^{L})\rightarrow P(\mathcal{A})$ maps the user query, local LLM's response and environment to a probability distribution over actions. A historical experience buffer $\mathcal{O}=(o_{0}, ..., o_{t}, ...,o_{N})$ records the past observations where $o_{t}=[(s_{t}, x, y^{L}), a_{t}]$ is the (user query, local LLM's response, environment)-action pair and $s_{i}\in\mathcal{S}$ is the environment state at time $t$. Inspired by the success of PPO \citep{schulman2017proximal}, we not only use the a reward function to evaluate current selected action but also leverage a critic function to estimate the cumulative value of historical action selections. Specifically, $\mathcal{R}$ can be denoted as $\mathcal{R}: (\mathcal{S}\times\mathcal{A})\rightarrow \mathbb{R}$ and $\mathcal{C}$ is $Q(s_{t}, a_{t})-V(s_t)$, where $Q(s, a)$ is the action-state value function and $V(s)$ is the state value function. $r_{t}=\mathcal{R}(s_t, a_t)$ can represent the reward received at time $t$. We'll further elaborate how these functions are used within $P^{3}Defer$ in the following sections. Our objective in this study is to learn the policy $\pi_{\theta}$ from given training dataset $\mathcal{D}$ to enable $\mathcal{A}$ being aware of both performance-cost and privacy concerns:
\begin{equation}
    \max_{\pi} \mathbb{E}_{\tau \sim \pi} \left[ \sum_{t=0}^{T} \gamma^t R_t \right]
\end{equation}
where \( \tau \) denotes a trajectory of state-action pairs.

\subsection{$\mathbf{P^{3}Defer}$}
Unlike existing methods where the deferral module is fixed and only makes decisions based on logits distribution or model's confidence level. The deferral module $D(\cdot)$ in $\mathbf{P^{3}Defer}$ is formulated as a reinforcement learning agent that selects among three actions: returning the local model's output $y^L$, requesting a response from the server LLM $y^S$, or masking private tokens before making a deferral decision. The module operates within an environment defined by the tuple $D(\mathcal{A}, \mathcal{R}, \mathcal{C}, \mathcal{S}, x, y^{L})$ which contains:\\
\textbf{Action Space ($\mathcal{A}$).} The available actions include:
\begin{itemize}
    \item $a_1$: accept $y^L$ and set final response $y=y^{L}$
    \item $a_2$: defer $x$ to $\Phi(S)$ and let final response $y=y^{S}$.
    \item $a_3$: apply privacy masking using private memory $\mathcal{M}$ on $x$ and routing the privacy masked $x'$ to $\Phi(S)$, final response $y=\Phi(S)(x')$.
\end{itemize}
\textbf{State Space ($\mathcal{S}$).} There are four environment states:
\begin{itemize}
    \item $x$ contains privacy concerns, $y^{L}$ is good.
    \item $x$ does not contain privacy concerns, $y^{L}$ is good.
    \item  $x$ contains privacy concerns, $y^{L}$ is bad.
    \item  $x$ does not contains privacy concerns, $y^{L}$ is bad.
\end{itemize}
Each state $s_t = [e_t^p, e_t^q]$ consists of privacy- and quality-related embeddings capturing the four possible states above based on the given input query $x$ and the local llm's response $y^L$.\\
\textbf{Reward Function ($\mathcal{R}$).} The reward function optimizes both response quality and privacy compliance:
\begin{equation}
    R_t = \mathbb{P}^{q}(y, \hat{y}) + \lambda \mathbb{P}^{p}(x)
    \label{eq:reward_function}
\end{equation}
where $\mathbb{P}^{q}(y, \hat{y})$ measures the generation quality of final response $y$ with respect to the golden responses $\hat{y}$, $\mathbb{P}^{p}(x)$ represents the identification of privacy leakage and $\lambda$ is a scaling factor. $\mathbb{P}^{q}$ and $\mathbb{P}^{p}$ are both in the form of entropy calculation as referred in Table \ref{tab:datasets}\\
\textbf{Critic Function ($\mathcal{C}$).} The critic model evaluates the expected return for different actions and guides the policy updates accordingly.
\begin{equation} 
    V^\pi(s_t) = \mathbb{E}{\pi} \left[ \sum{t'=t}^{T} \gamma^{t'-t} R_{t'} \right]
\end{equation}
where \( \gamma \in [0,1] \) is the discount factor, controlling the importance of future rewards; \( T \) is the trajectory length; \( R_t \) is the immediate reward as defined in your equation. The expectation is taken over all possible trajectories following policy $\pi_\theta(a|s)$.\par
Additionally, the state-action value function (Q-function) is:
\begin{equation}
    Q^\pi(s_t, a_t) = \mathbb{E}_{\pi} \left[ \sum_{t'=t}^{T} \gamma^{t'-t} R_{t'} \mid s_t, a_t \right]
\end{equation}
which evaluates the expected reward after taking action \( a_t \) in state \( s_t \).\par
Using the PPO framework, the policy network $\pi_\theta(a|s)$ updates its parameters $\theta$ by maximizing the following objective:
\begin{equation}
    \small
    \nabla_{\theta} J(\pi_{\theta}) = \mathbb{E}_{\tau \sim \pi_{\theta}} \left[ \sum_{t=0}^{T} \nabla_{\theta} \log \pi_{\theta}(a_t \mid s_t, x, y^L) \hat{A}_t \right]
\end{equation}
where $\hat{A}_t =Q^\pi(s_t, a_t) - V_{\psi}(s_t)$ is the advantage function.

\subsection{Local LLM Training}
The local LLM \( \Phi(L) \) is trained using two key techniques: (1) \textbf{CoT-enhanced instruction tuning} and (2) \textbf{knowledge distillation} from the server LLM \( \Phi(S) \) when deferral occurs.

\textbf{CoT-enhanced Instruction Tuning}
To improve reasoning capabilities, we fine-tune the local LLM using a dataset of instruction-response pairs enhanced with chain-of-thought (CoT) reasoning. In this paper, we mainly use the zero-shot CoT prompting to formulate our instructions as can be seen in appendix \ref{sec:appendix_prompts}. Given an instruction \( x \) and the corresponding target response \( \hat{y} \), the loss function is defined as:

\begin{equation}
    \mathcal{L}_{\text{inst}} = - \sum_{t} \log P_{\Phi(L)}(\hat{y}_t \mid x, \hat{y}_{<t}).
\end{equation}

This objective encourages the model to generate responses aligned with human-annotated outputs while incorporating reasoning steps.

\textbf{Knowledge Distillation from Server LLM}
When the server LLM \( \Phi(S) \) is invoked due to deferral, the local LLM learns from the distilled server responses \( y^S \). The knowledge distillation loss minimizes the divergence between the local and server predictions:

\begin{equation}
    \small
    \mathcal{L}_{\text{KD}} = \sum_{t} D_{\text{KL}} \left( P_{\Phi(S)}(y^S_t \mid x, y^S_{<t}) \parallel P_{\Phi(L)}(y^S_t \mid x, y^S_{<t}) \right),
\end{equation}

where \( D_{\text{KL}} \) is the Kullback-Leibler divergence. This loss ensures that the local model mimics the server model’s outputs when necessary.

\textbf{Training Objective}
The overall training objective combines the two losses:

\begin{equation}
    \mathcal{L} = \mathcal{L}_{\text{inst}} + \lambda_{\text{KD}} \mathcal{L}_{\text{KD}},
\end{equation}

where \( \lambda_{\text{KD}} \) controls the influence of knowledge distillation. This framework enables the local LLM to improve its reasoning and generalization capabilities while reducing reliance on the server.

\subsection{Private Memory $\mathcal{M}$}
Unlike previous work \citep{hartmann2024can}, which relies on the local LLM to identify and rewrite private tokens, this approach carries the risk of altering the original meaning of the given query during the rewriting process. To address this issue, we introduce an innovative private memory \(\mathcal{M}\) that pre-stores private tokens extracted from a large corpus \citep{zhang-etal-2024-llm-based}. When a private query is encountered, the private memory efficiently identifies and masks private tokens without modifying the original meaning. \\
The memory is structured as a dynamic, growing list, where private tokens are detected by measuring the Levenshtein distance of each token. Once identified, replacing private tokens with similar alternatives helps mitigate privacy leakage while preserving the original intent of the query, thereby ensuring the quality of the final response.

\subsection{Inference}
\begin{figure}[h]
    \centering
    \includegraphics[width=0.5\textwidth]{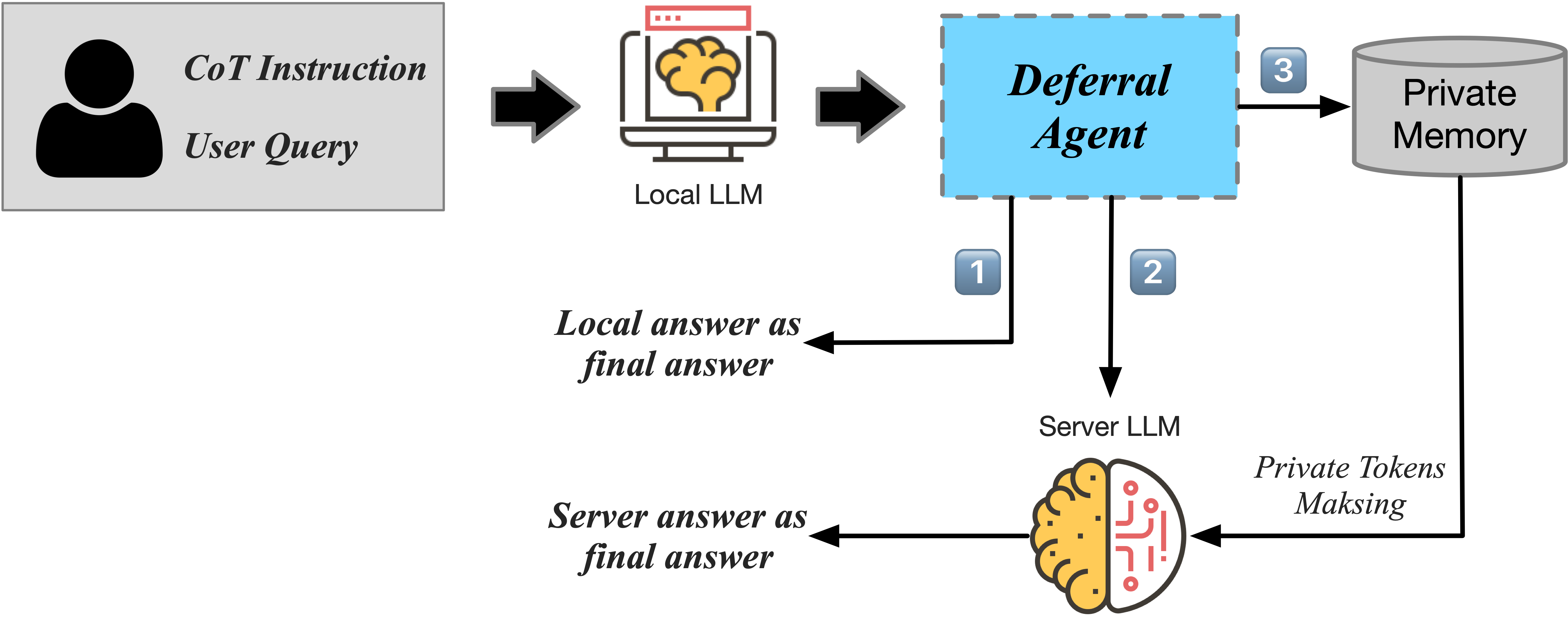}
    \caption{Inference process of the proposed framework. The \textit{Deferral Agent} determines whether to return the local response, defer to the server LLM, or apply privacy masking based on the input query.}
    \label{fig:inference}
\end{figure}

During inference, the user query \( x \) is first processed by the local LLM \( \Phi(L) \), generating an initial response \( y^L \). The \textit{Deferral Agent} $\mathcal{A}$ then decides among three actions based on the query context: \textit{Use local response:} If the local LLM’s response is deemed sufficient, it is returned as the final answer; \textit{Defer to server LLM:} If the query is too complex or uncertain and presents no privacy concerns, the agent queries the server LLM \( \Phi(S) \), which provides a refined response \( y^S \); \textit{Mask private tokens:} If sensitive information is detected, the agent applies a privacy-preserving mechanism using the \textit{Private Memory} module before passing the modified query to the server. The agent optimally selects actions to balance response quality and privacy, ensuring reliable and secure query resolution.

\section{Experimental Settings}
\subsection{Datasets}
To validate the effectiveness of $P^{3}Defer$, we opt for three benchmarks with privacy concerns that cover daily scenarios in on-device intelligence application to test our methods as below, more statistics can be seen in appendix \ref{sec:appendix_supplementary_datasets} and Table \ref{tab:datasets}. \par
\textbf{GSM8K}\citep{cobbe2021training} is a graduate student mathematical dataset consisting of mathematical questions and corresponding solutions, of which some questions contain personal information for privacy study\citep{hartmann2024can}. \par
\textbf{MedSum}\citep{zekaoui2023enhancing} is a medical related dataset with a focus on summarizing the customer health question. The dataset contains customer health questions and corresponding summaries which contains personal healthcare information. \par
\textbf{EmailSum}\citep{zhang2021emailsum} is a sequence-to-sequence email summarization dataset consisting of daily email thread and corresponding summary. The summary types are available in long-summary and short-summary, we use short-summary in this study. \par

\subsection{Tasks, Metrics \& Baselines}
We evaluate our proposed $P^{3}Defer$ on three commonly used daily tasks: mathematical QA, medical inquiry summarization, and email summarization, as indicated in Table \ref{tab:datasets}. We also incorporate the metric of privacy leakage \citep{hartmann2024can}, which calculates the average number of privacy tokens leaked when sending queries to the server LLM (Check more details in appendix \ref{sec:appendix_supplementary_datasets}). \par
To the best our knowledge, rare study has been made in privacy-preserved LLM cascade except \citet{hartmann2024can} leverages in-context learning for query rewritten to mitigate privacy leakage problem. Thus, we first compare our $P^{3}Defer$ with existing logit-based \citep{wang2024cascade, jitkrittum2024does} and confidence-based \citep{zhu-etal-2024-towards} cascade methods. For logit-based methods, we are using Instruction Tuning (IT) and Loss Tuning (LT); for confidence-based method, we are using Few-shot In-context Learning (Few-shot ICL), detailed implementation can be seen in appendix \ref{sec:appendix_methodology}. Further, we compare our $P^{3}Defer$ with other policy learning methods that close to our work: TREACLE\citep{zhangefficient} and Bilevel\citep{yan2023ask}. Next, we conduct privacy study to evaluate how $P^{3}Defer$ mitigates privacy problem and compare our work with \citet{hartmann2024can}. Extensive experiments validate the efficiency and superiority of $P^{3}Defer$ in privacy-preserved LLM cascade.

\subsection{Implementation Details}
For implementation details, we leverage the Transformers\citep{wolf-etal-2020-transformers} as the base code and conduct extensive experiments with the Gemma models\citep{team2024gemma}: \textbf{Gemma-2B} as the local LLM, \textbf{Gemma-7B} as the server LLM. Notably, the server LLM is fine-tuned on all datasets to reach reasonably great performance, of which the server LLM's ability on GSM8K, MedSum and EmailSum are 52.85\%, 61.22\% and 56.51\%, respectively. We use the AdamW optimizer\citep{loshchilov2018decoupled, paszke2017automatic} with a learning rate of 5e-4 and also a linear warm-up scheduler initialized with 10\% of the total training steps as warm-up steps and a weight decay of 1e-4 to avoid over-fitting for all the experiments. The batch size per device is set to 8. All the experiments are conducted on two computation nodes configured with eight 80G H100 GPUs.

\section{Experimental Results}
\subsection{Cascade Study}
\begin{table*}[ht]
    \centering
    \small
    \begin{tabular}{cccccccc}
        \hline
         \multicolumn{3}{c}{Method Type}& Confidence-based & \multicolumn{2}{c}{Logit-based} & \multicolumn{2}{c}{Policy Learning} \\
         Dataset & Metric & \% & Few-shot ICL & IT & LT & TREACLE & P$^{3}$Defer \\
         \hline
         \multirow{5}{*}{GSM8K} & CR & & 100 & 100 & 81.2 & 93.1 & {\color{red}\textbf{66.41}} \\
           & SCR & & 28.13 & 28.13 & 31.75 & 84.31 & {\color{blue}\textbf{92.61}} \\
           & \multirow{3}{*}{Acc} & $\Phi(L)$ & 11.83 & 26.08 & 26.91 & 24.31 & 27.33 \\
           & & $\Phi(L)+\Phi(S)$ & 52.85 & 52.85 & 55.92 & 55.78 & {\color{red}\textbf{55.96}} \\
           & & vs $\Phi(S)$ & N.A. & N.A. & $\uparrow$3.07 & $\uparrow$2.07 & $\uparrow$\textbf{3.11} \\
         \hline
         \multirow{5}{*}{MedSum} & CR & & 96.2 & 94.8 & 97.3 & 80.6 & {\color{red}\textbf{69.71}} \\
           & SCR & & 26.09 & 26.89 & 26.92 & 76.93 & {\color{blue}\textbf{88.40}} \\
           & \multirow{3}{*}{R-S} & $\Phi(L)$ & 28.55 & 34.61 & 36.77 & 34.87 & 35.31 \\
           & & $\Phi(L)+\Phi(S)$ & 61.97 & 62.18 & 62.95 & 63.17 & {\color{red}\textbf{63.94}} \\
           & & vs $\Phi(S)$ & $\uparrow$0.75 & $\uparrow$0.96 & $\uparrow$1.73 & $\uparrow$1.95 & $\uparrow$\textbf{2.72} \\
         \hline
         \multirow{5}{*}{EmailSum} & CR & & 100 & 98.5 & 80.6 & 88.9 & {\color{red}\textbf{44.7}} \\
           & SCR & & 31.77 & 39.16 & 46.93 & 79.16 & {\color{blue}\textbf{94.61}} \\
           & \multirow{3}{*}{R-S} & $\Phi(L)$ & 24.59 & 29.49 & 28.58 & 27.06 & 28.91 \\
           & & $\Phi(L)+\Phi(S)$ & 56.51 & 56.92 & 56.99 & 60.19 & {\color{red}\textbf{61.21}} \\
           & & vs $\Phi(S)$ & N.A. & $\uparrow$0.41 & $\uparrow$0.48  & $\uparrow$3.68 & $\uparrow$\textbf{4.70} \\
         \hline
    \end{tabular}
    \caption{The best cascade performance of $\Phi(L)$ across three benchmarks. CR denotes call rate, indicating the proportion of queries sent to the server. SCR represents safe call rate, reflecting the number of queries that are safe (i.e., those sent to the server that do not contain privacy information) among the total sent queries. Acc refers to accuracy, while R-S indicates the ROUGE-Sum score. The symbol $\uparrow$ signifies an improvement compared to $\Phi(S)$. The {\color{red}\textbf{red}} number pair shows the best cascade performance (lower call rate with higher scores), the {\color{blue}\textbf{blue}} number indicates the safest method.}
    \label{tab:cascade_results}
\end{table*}

\begin{figure*}[ht]
    \centering
    \subfigure[GSM8K]{\includegraphics[width=0.3\textwidth]{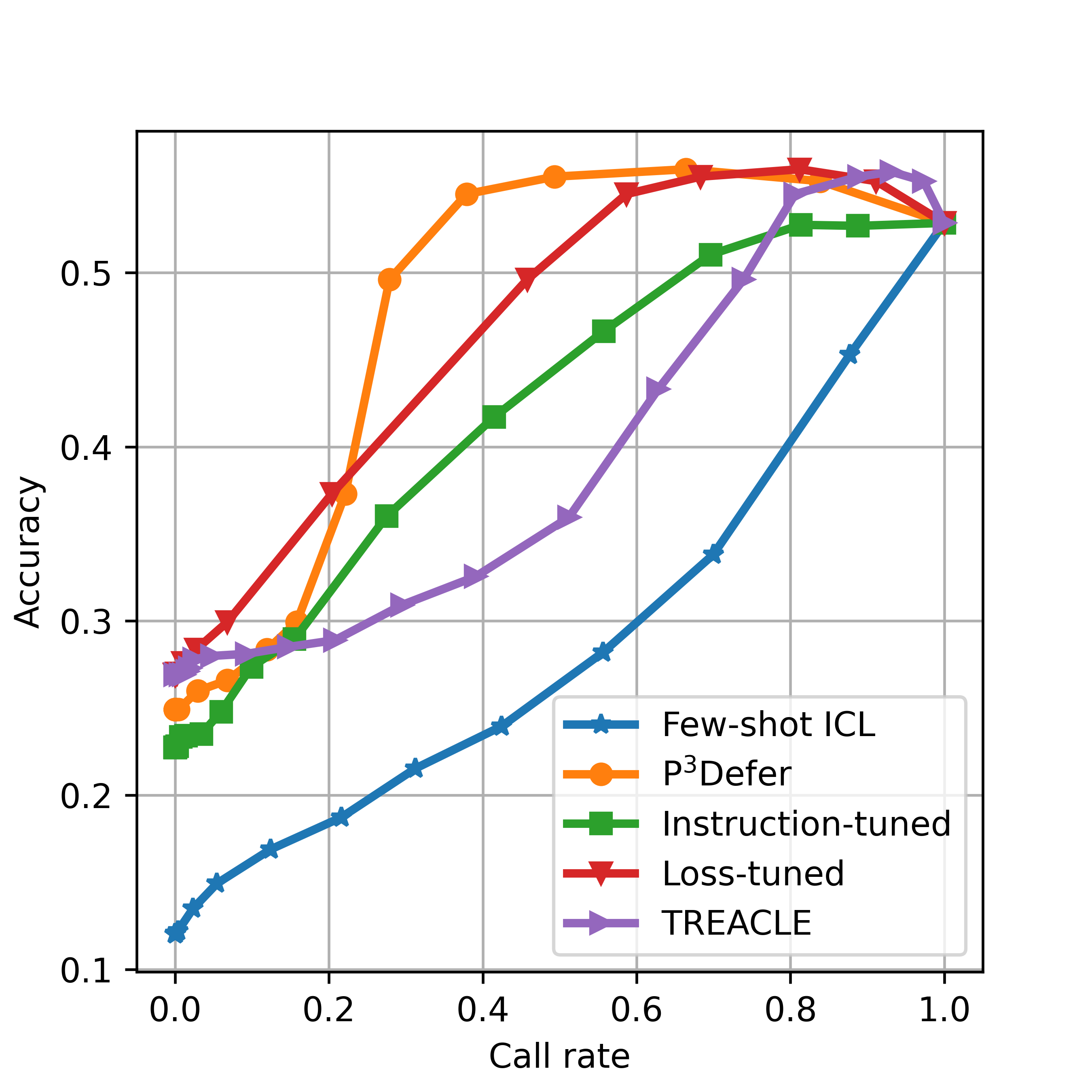}} 
    \subfigure[MedSum]{\includegraphics[width=0.3\textwidth]{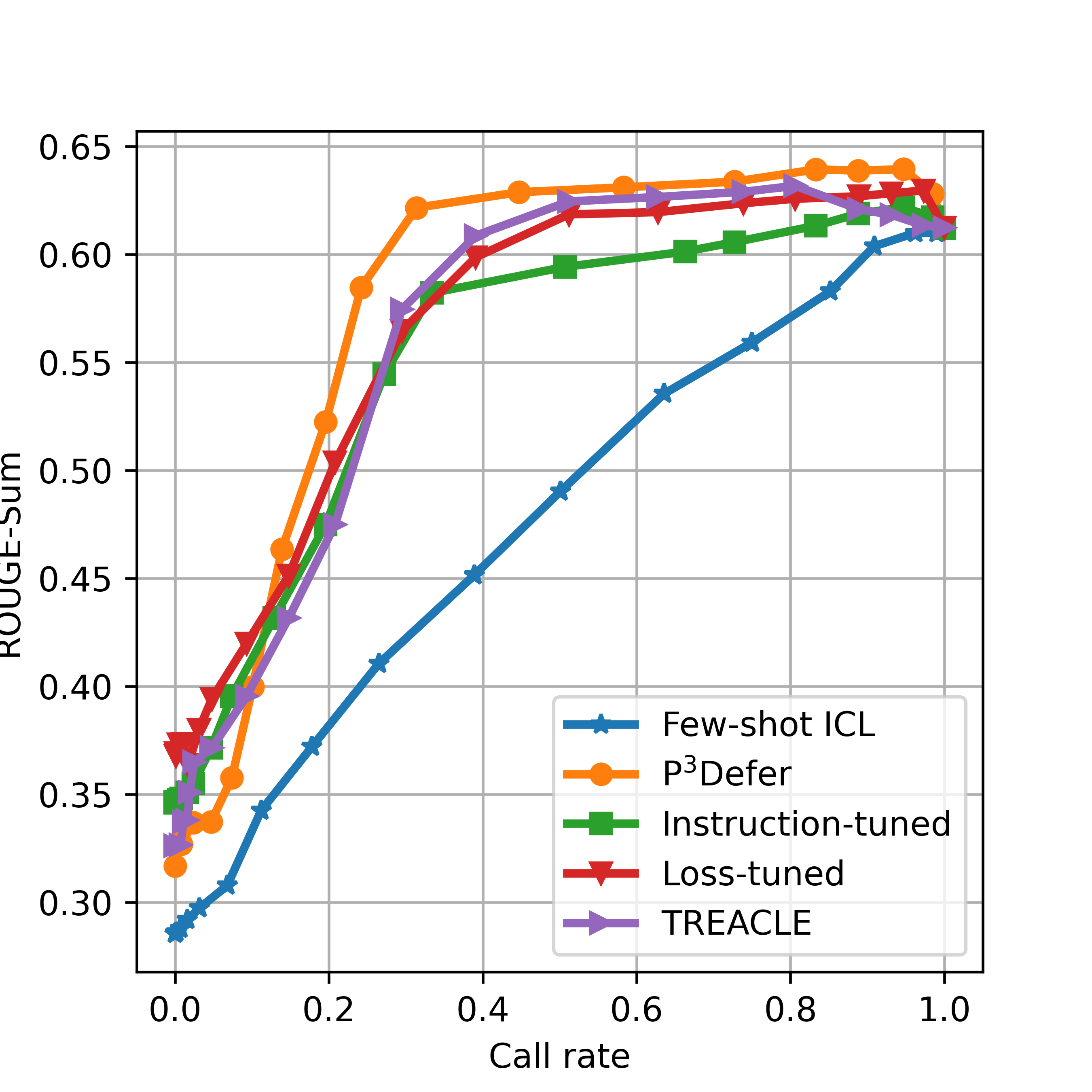}} 
    \subfigure[EmailSum]{\includegraphics[width=0.3\textwidth]{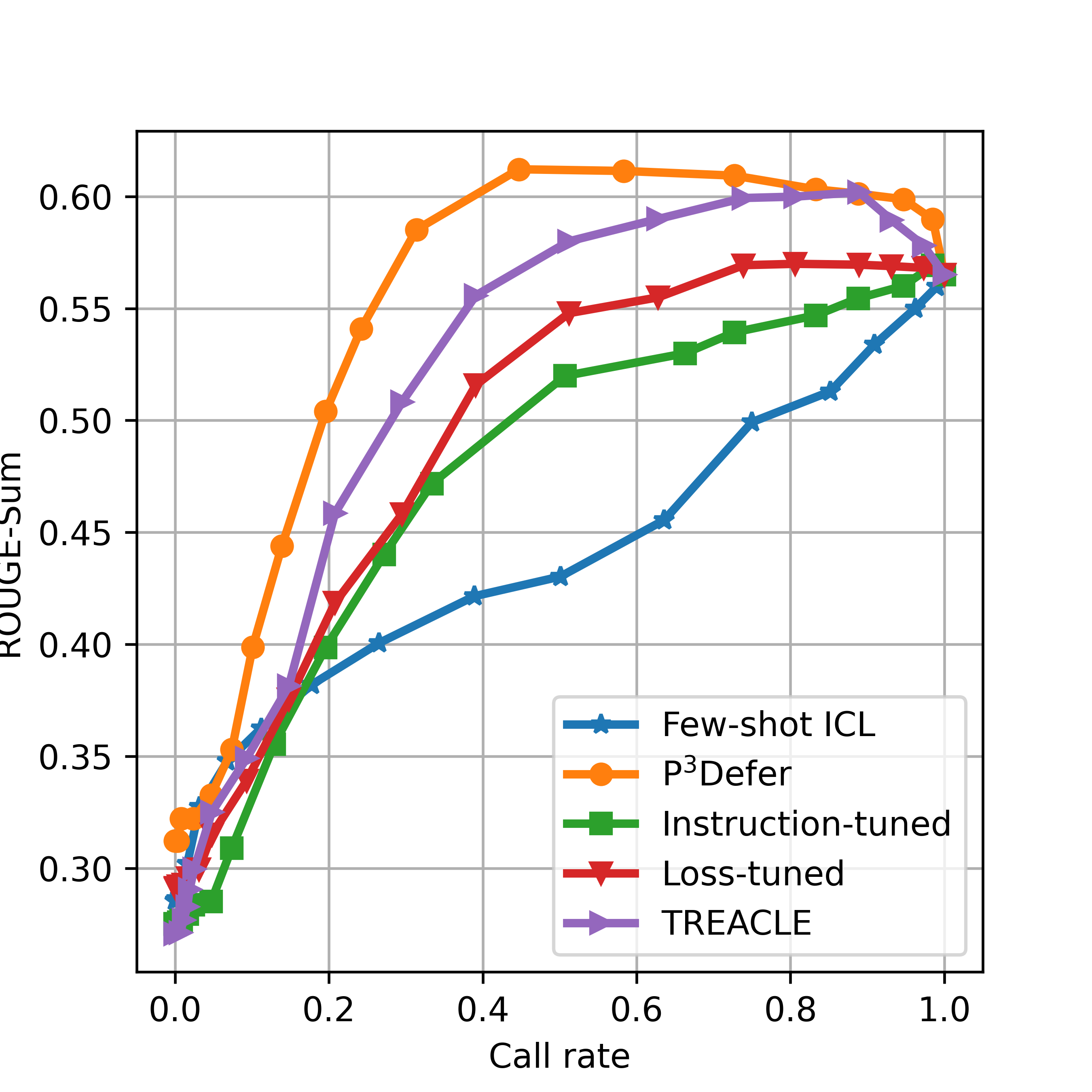}}
    \caption{Curves depicting cascade performance versus call rate for different methods across all three datasets: (a) GSM8K, (b) MedSum, and (c) EmailSum.}
    \label{fig:cascade_figure}
\end{figure*}

\textbf{Cascade Performance} One of the key advantages of LLM cascading is its ability to enhance performance without increasing the size of the base local LLM. As shown in Table \ref{tab:cascade_results}, confidence-based models primarily rely on the server LLM to boost performance, while logit-based methods selectively defer difficult queries that the local model cannot solve, leading to performance improvements. In contrast, our proposed $P^{3}Defer$ achieves state-of-the-art performance across all three benchmarks, demonstrating an accuracy of 55.96\% with a call rate of 66.41\% on GSM8K, a ROUGE-Sum score of 63.94\% with a call rate of 69.71\% on MedSum, and a ROUGE-Sum score of 61.21\% with a call rate of 44.7\% on EmailSum. Notably, $P^{3}Defer$ outperforms all other baselines, achieving post-cascade improvements of 3.11\%, 2.72\%, and 4.70\% over the server model across the three datasets, respectively. \\
\textbf{Performance vs Cost} A crucial factor in evaluating an LLM cascade system is the trade-off between performance and cost, where the ideal approach maximizes performance gains while minimizing the server call rate. As observed in Table \ref{tab:cascade_results}, policy learning-based methods, such as TREACLE and $P^{3}Defer$, make fewer calls to the server while still improving performance, distinguishing them from confidence-based and logit-based approaches. Furthermore, as depicted in Figure \ref{fig:cascade_figure}, $P^{3}Defer$ demonstrates superior deferral decision-making, as its performance curve reaches an inflection point earlier while attaining the highest performance compared to other methods. Moreover, in Figure \ref{fig:cascade_figure}, we observe that TREACLE exhibits different trends on the GSM8K dataset compared to the two summarization datasets. We attribute this to TREACLE’s reliance on its routing strategy rather than enhancing the local LLM’s capabilities, whereas other methods focus on both cascade deferral and improving the local LLM. Additionally, an interesting observation is that the confidence-based method demonstrates inconsistencies across the three datasets, suggesting that instructing the local LLM for cascading leads to unreliable performance. These findings highlight the effectiveness and superiority of $P^{3}Defer$ in optimizing cascade performance while maintaining cost-efficient.

\subsection{Privacy Study}
\begin{table*}[ht]
    \centering
    \small
    \begin{tabular}{ccccccc}
        \hline
         Dataset & Metric & Few-shot ICL & Instruction Tuning & Loss Tuning & TREACLE & P$^{3}$Defer \\
        \hline
         \multirow{3}{*}{GSM8K} & precision & 64.17 & 82.95 & 91.79 & 88.17 & \textbf{96.31} \\
           & recall & 44.20 & 72.89 & 87.24 & 76.45 & \textbf{88.79} \\
           & r(leakage) & 95.11 & 84.17 & 75.98 & 74.22 & \textbf{20.11} \\
        \hline
         \multirow{3}{*}{MedSum} & precision & 68.85 & 85.62 & 90.10 & 87.41 & \textbf{92.17} \\
           & recall & 42.99 & 68.84 & 82.99 & 68.41 & \textbf{88.56} \\
           & r(leakage) & 97.60 & 72.14 & 70.10 & 70.54 & \textbf{23.87} \\
        \hline
        \multirow{3}{*}{EmailSum} & precision & 68.85 & 85.62 & 90.10 & 82.17 & \textbf{96.91} \\
           & recall & 42.99 & 68.84 & 82.99 & 62.43 & \textbf{85.77} \\
           & r(leakage) & 80.79 & 73.46 & 56.52 & 55.62 & \textbf{16.34} \\
        \hline
    \end{tabular}
    \caption{Privacy study. Precision and recall are used for evaluating the ability of different methods on identifying queries with privacy concerns, r(leakage) measures the ratio between leaked private tokens and all private tokens.}
    \vspace{-10pt}
    \label{tab:privacy}
\end{table*}

Beyond its improvements in cascade performance, our $P^{3}Defer$ also demonstrates a remarkable ability to mitigate privacy concerns. As shown in Table \ref{tab:cascade_results}, $P^{3}Defer$ achieves a safe call rate of 92.61\%, 88.40\%, and 94.61\% across the three datasets, respectively. Notably, confidence-based methods achieve only around 28.66\%, indicating that relying solely on the local LLM to identify privacy-sensitive queries is unreliable. Moreover, while logit-based methods offer some improvements in privacy sensitivity, they still fall short compared to policy-learning-based approaches. This finding is further validated by the results in Table \ref{tab:privacy}, where the precision and recall scores of confidence- and logit-based methods remain inferior to those of policy-learning-based methods. Similar patterns emerge in mitigating privacy leakage, as confidence-based methods leak the most private tokens across all three datasets, reinforcing the unreliability of instructing the model itself to rewrite queries. Although logit-based methods provide some mitigation, their performance remains suboptimal. We attribute this to the fundamental limitation of logit-based methods: their primary objective is to align logits with quality confidence, making them unsuitable for incorporating additional considerations such as privacy protection during deferral decisions. In contrast, $P^{3}Defer$ achieves average relative reductions of 75.35\%, 68.64\%, and 74.81\% in leaked token ratios across the three datasets. This substantial reduction highlights the advantages of $P^{3}Defer$ in handling privacy-sensitive queries, which we attribute to the integration of private memory. By leveraging private memory that pre-stores private tokens, the local LLM does not need to focus on rewriting queries. Instead, the memory mechanism assists in identifying and masking private tokens before sending queries to the server, leading to significant improvements in mitigating privacy leakage. Together, policy learning enables $P^{3}Defer$ to accurately identify privacy-sensitive queries, while private memory effectively mitigates private token leakage, ensuring a more secure and privacy-aware LLM cascade system.

\subsection{Ablation Study}
\textbf{Ablation on Cascade Performance}
\begin{figure}[ht]
    \centering
    \includegraphics[width=0.8\linewidth]{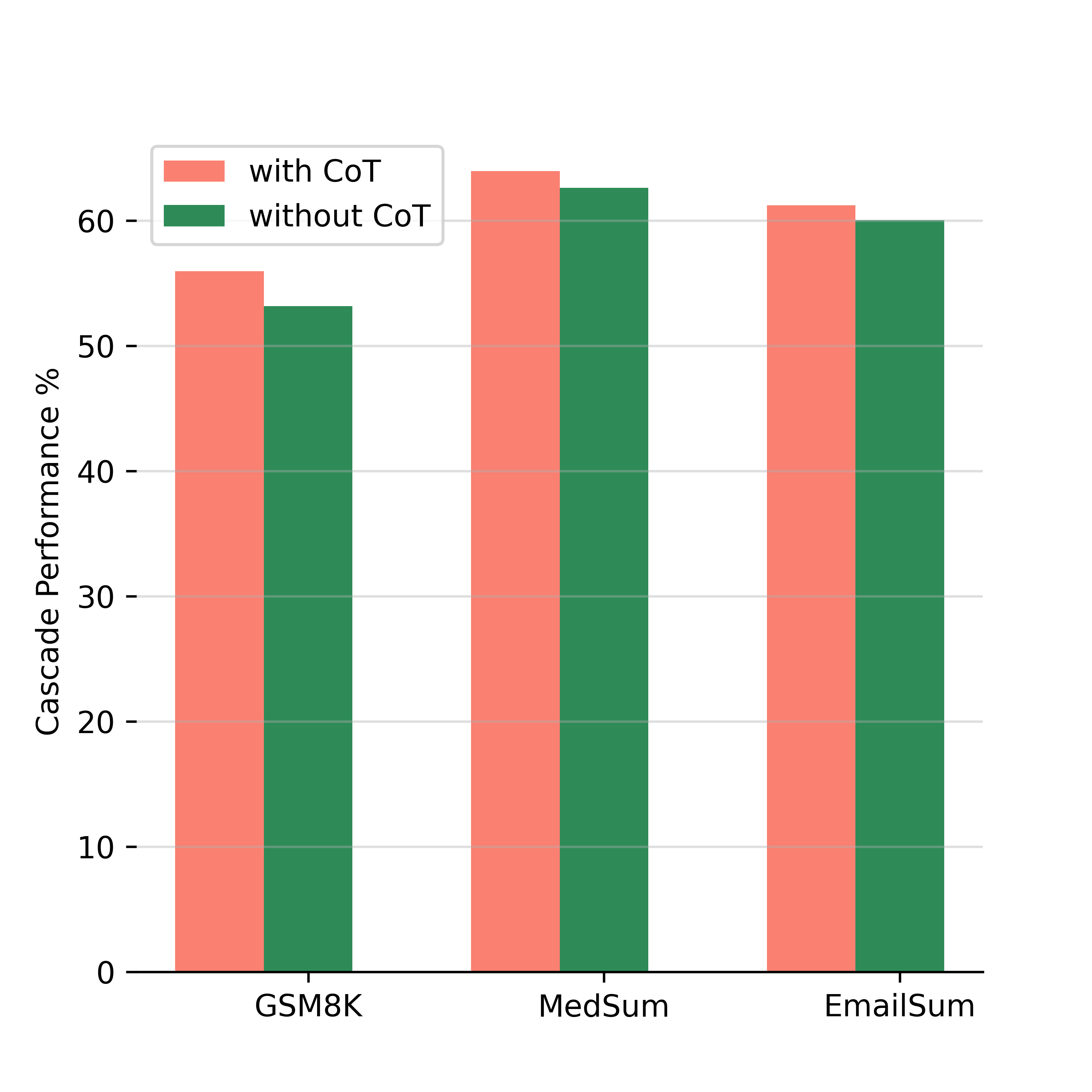}
    \caption{Ablation study on CoT usage.}
    \label{fig:ablation_cot}
\end{figure}
We further conduct ablation study on the usage of CoT. The results are presented in Figure \ref{fig:ablation_cot}, we observe that incorporating CoT reasoning consistently improves cascade performance across all datasets, albeit with varying magnitudes. The most significant improvement is observed on the GSM8K dataset, where the model with CoT outperforms its counterpart without CoT by approximately 3\%. This suggests that CoT reasoning enhances logical reasoning capabilities, allowing the local model to make better-informed cascade decisions. For MedSum and EmailSum, the performance gap between CoT and non-CoT models is relatively smaller (around 1-2\%) which we attribute to the fact that MedSum and EmailSum rely more on semantic understanding and less on multi-step reasoning, making CoT less critical in these cases. Overall, these findings suggest that CoT is a beneficial augmentation to local model training, particularly in reasoning-intensive tasks which further validate the effectiveness of the whole $P^{3}Defer$ design.
\textbf{Ablation on Privacy Preservation}
\begin{figure}[ht]
    \centering
    \includegraphics[width=0.8\linewidth]{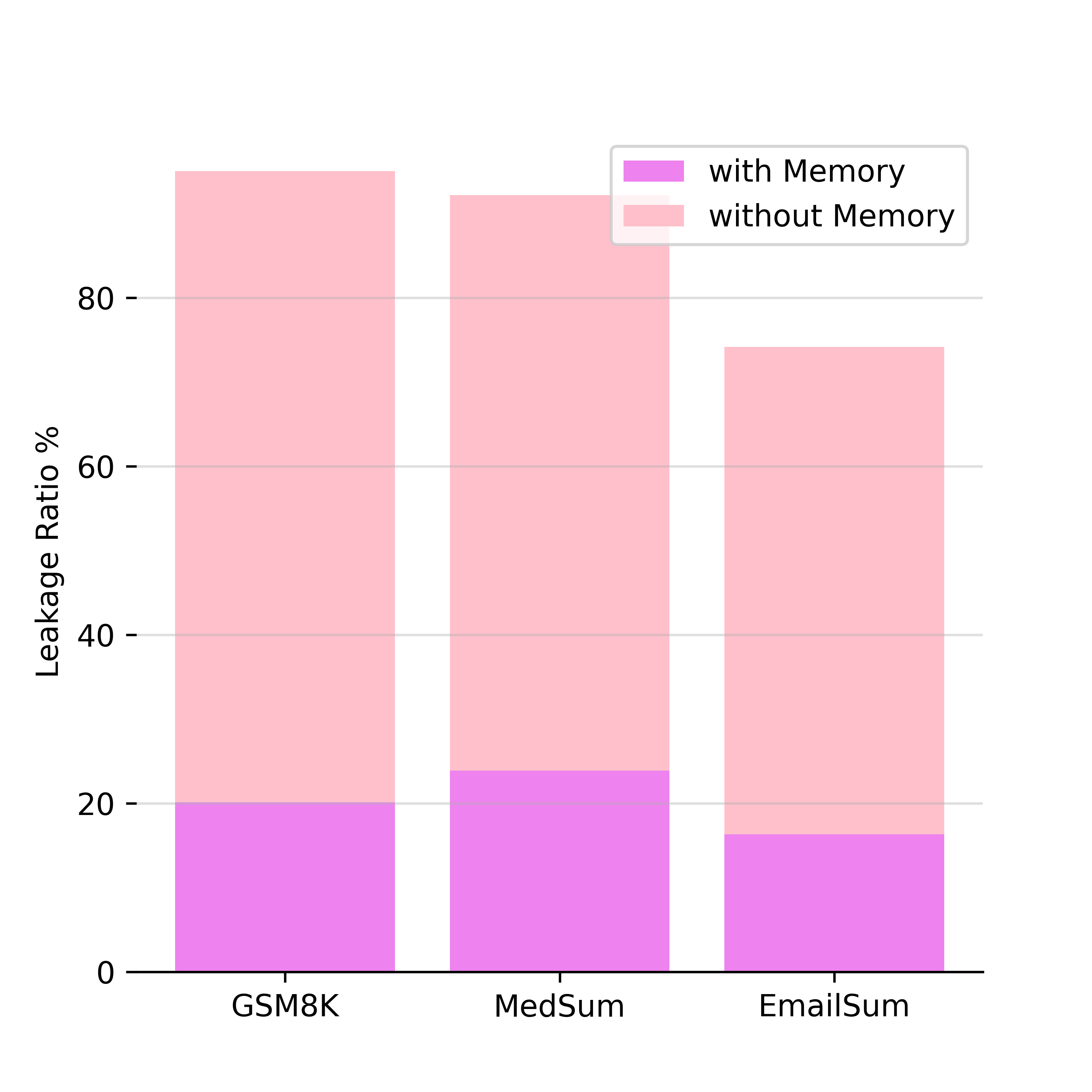}
    \caption{Ablation study on private memory usage.}
    \label{fig:ablation_memory}
\end{figure}
Beyond cascade performance, privacy preservation is also a crucial objective of our approach. To further investigate the impact of memory design, we conduct an ablation study on the memory component, as shown in Fig. \ref{fig:ablation_memory}. The results reveal that leveraging private memory significantly mitigates privacy token leakage, as indicated by the substantially lower violet bar compared to the pink one. This demonstrates that private memory is far more effective than relying on the local LLM’s rewriting ability to reduce private token leakage, further validating the overall design of $P^{3}Defer$ of which, policy learning endorse $P^{3}Defer$ the ability to accurately detect privacy-sensitive queries while private memory serves for mitigates private token leakage, ensuring an effective and privacy-preserved LLM cascade system.

\section{Conclusion \& Furture Work}
In this study, we advance the privacy-preserved LLM cascade by incorporating policy learning coupled with a private memory, moving beyond existing approaches that primarily emphasize cost-performance trade-offs. This enhancement aligns more closely with the demands of real-world applications. Extensive experiments demonstrate that $P^{3}Defer$ significantly mitigate the privacy leakage problem while improving the llm cascade systemperformance.\par
While this work represents the pioneer effort to introduce privacy-preserved LLM cascade, future research will explore more on other factors that fit real-world cascade system. We also aim to develop more computational efficient and multi-objective optimized methods to sustain favorable cost-performance trade-offs while accommodating a wider array of objectives such as latency. Innovations on training local llm and deferral module together are also worth to investigate.

\section{Related Work}
\textbf{LLM Cascade} Cascading has been extensively studied and applied across various domains due to its ability to enhance system performance in downstream tasks by selecting appropriate models \citep{hu2023cascaded, li2019rethinking, karlos2016semisupervised, viola2001rapid}. Recently, this approach has garnered increasing attention for improving the performance of large language models (LLMs). For instance, \citet{agrawal2024vidur, xu2023llmcad, chen2024model} have explored speculative decoding, which leverages a larger and more powerful LLM to verify token-level accuracy during the inference of a smaller LLM, thereby accelerating the overall process. Despite the success of cascading, researchers have observed that larger, more capable LLMs (e.g., GPT-4 \citep{achiam2023gpt}) can be expensive, while smaller LLMs (e.g., GPT-2 \citep{radford2019language}) may not always meet performance requirements. This has led to the emergence of the deferral rule—determining when to invoke the larger LLM—as a critical area of exploration for balancing performance and cost in LLM cascading \citep{shekhar2024towards, chen2023frugalgpt, chen2023less}. There are two primary approaches to deferral: confidence-based methods and router-based methods. Confidence-based methods leverage the LLM's confidence in its generated answers to inform deferral decisions. Ideally, an LLM exhibits higher confidence for higher-quality answers, and vice versa. A straightforward approach involves asking the LLM to provide a confidence score alongside its answers, invoking the stronger LLM when the score is low \citep{zhu-etal-2024-towards}. Another prevalent method utilizes the logits of generated tokens to represent the LLM's confidence, with recent studies exploring operations on logits, such as mean \citep{gupta2024language} and quantile \citep{jitkrittum2024does}. \citet{wang2024cascade} extended this concept by incorporating the logits of the stronger LLM into the loss function for tuning the weaker LLM, enhancing its understanding of the cascade logic and enabling deferral decisions based on logit indicators. In contrast, router-based methods use a routing mechanism to determine whether to invoke the stronger LLM. Typically, the router selects the most suitable LLM for different tasks. Non-predictive routing evaluates the outputs of multiple LLMs to select the best one, but this can be costly due to the need to assess all involved models \citep{madaan2023automix, lee2023orchestrallm, wang2023tabi}. Predictive routing systems, however, employ reward functions that allow the router to anticipate which LLM to select, thus avoiding the latency associated with extensive evaluations \citep{shnitzer2023large, vsakota2024fly, hari2023tryage}. Nonetheless, router-based methods require prior knowledge of each LLM's capabilities and may incur significant costs when trying to enhance performance, compared to confidence-based methods \citep{hu2024routerbench, hu2024mars}. Different from existing methods, we incorporate a CoT-enhanced policy learning strategy coupled with a private memory design to achieve privacy-preserved LLM cascade.

\textbf{Privacy-preservation} Privacy has always been a core concern in LLM research \citep{kim2024propile, zhang2024no, das2024security, janryd2024preventing, feng2024exposing}, particularly for on-device LLM applications \citep{zhang2024enabling, peng2024pocketllm, yuan2024wip}. LLMs have been shown to inadvertently reveal sensitive information, such as personal names \citep{evertz2024whispers, kim2024propile}. To address these privacy issues, \citet{liu2024rethinking, liu2024learning, liu2024towards, kassem2023preserving} proposed machine unlearning techniques that enable LLMs to forget sensitive information, thus mitigating the risk of generating harmful or biased content. Another approach is differential privacy, which adds noise to the training data, making it more difficult to identify individual data points \citep{hartmann2024can}. Additionally, \citet{zhang2024get} suggested using contrastive learning to erase an LLM’s memory of user information. While these methods have shown success across diverse user bases, our objective is to enhance the sensitivity of our LLM cascade framework to privacy concerns in single-user settings. To achieve this, we aim to leverage in-context learning and integrate binary privacy identification into the loss function, allowing the local LLM to be more attuned to privacy considerations during the cascading process. Further, we innovatively utilize a private memory into our design to achieve privacy-preseveration.

\section*{Limitations}
Despite the empirical success, our $P^{3}Defer$ presents two limitation that may ask for further attentions to work on. \textbf{First}, compare with confidence- and logit-based methods that leverage thresholds to make deferral decisions, our method needs to train a policy that contains four components (even some of them have small set of parameters), the computational costs are higher. However, the higher costs obtain a reasonable feedback on the performance and privacy-preservation ability. We may still want to seek ways for reducing the computational costs\citep{zhou2023offline}. \textbf{Second}, our private memory design is a pre-process which means it can not be updated even new privacy tokens appear. This may pose hackers a way to attack this system by simply use synonym\citep{zhang-etal-2024-llm-based}. Further explorations in including other memory techniques\citep{zhang2024personalized} can be important.

\section*{Ethics Statement}
After carefully reviewing the ACL Ethics Policy, we are committed to show our respect and obey to consent all.

\section*{Acknowledgment}
Thanks for ACL community and Openreview platform, will further add details in the final revision. 

\bibliography{anthology,custom}
\bibliographystyle{acl_natbib}

\appendix
\section{Prompts}
\label{sec:appendix_prompts}
\begin{figure}[ht]
    \centering
    \includegraphics[width=0.4\textwidth]{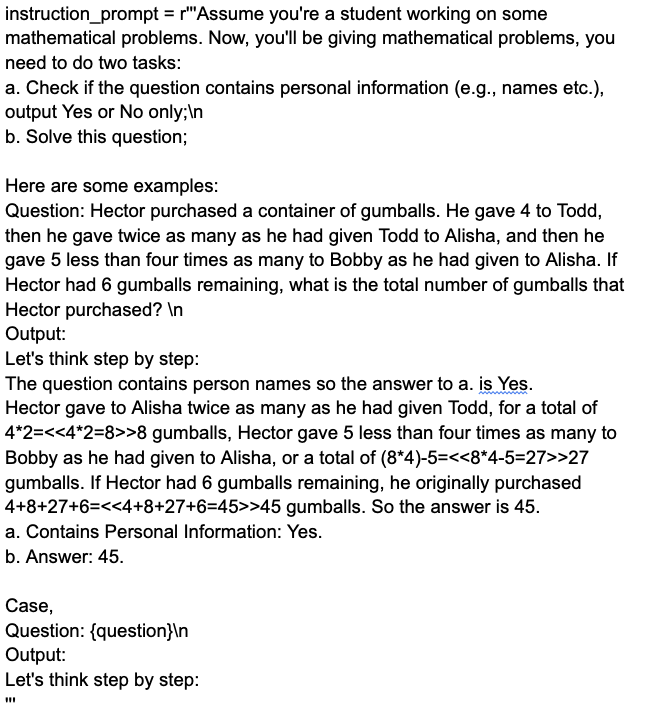}
    \caption{Prompts Used on three datasets.}
    \label{fig:prompts}
\end{figure}

The design of prompts plays a crucial role in activating the LLM's capabilities for downstream tasks. Following the findings of \citet{webson2021prompt} on prompt design, we first assume a persona for the LLM, then provide task instructions and ask the model to generate outputs in a fixed style. For few-shot prompting, we provide task examples along with their corresponding outputs; details are shown in Figure \ref{fig:prompts}. Interestingly, we observed that as the number and complexity of tasks in the instructions increased, the model's performance on the target task declined, as demonstrated in Table \ref{tab:cascade_results}. The prompts presented here yielded the best performance among all the variations we tested. \\

\section{Preliminary Results}
\label{sec:appendix_preliminary}
\begin{table*}[ht]
    \centering
    \small
    \begin{tabular}{ccccccc}
        \hline
         \multirow{2}{*}{Metric \%} & \multirow{2}{*}{Cascade} & \multicolumn{4}{c}{Prompt Engineering} & \multirow{2}{*}{Instruction Tuning} \\
         & & 0-shot & 1-shot & 2-shot & 5-shot & \\
        \hline
         Call Rate & & 0 & 70.43 & 48.98 & 67.43 & 42.76  \\
         Safe Call Rate & & 0 & 2.05 & 2.94 & 2.13 & \textbf{27.61}  \\
         \multirow{2}{*}{Accuracy} & \XSolidBrush & 14.94 & 10.08 & \textbf{11.83} & 10.68 & \textbf{26.08}  \\
         & \Checkmark & 14.94 & 42.91 & 37.30 & 42.61 & 42.29 \\
        \hline
    \end{tabular}
    \caption{Preliminary results on GSM8K.}
    \label{tab:preliminary_results}
\end{table*}
Following the approach of \citet{hartmann2024can}, we initially attempted to use self-critique and rely on the in-context learning capabilities of the local LLM to implement the deferral function. Specifically, we instructed the model to handle the task while simultaneously outputting a confidence level, which would determine whether the query should be deferred to the server. However, preliminary results revealed limitations in this design. As shown in Table \ref{tab:preliminary_results}, without examples, the local model tends to be overly confident in every generated response. Moreover, even when provided with several examples, the model treats confidence as a classification task, rather than correlating it with the quality of its generated responses. Consequently, we opted to use logits for more effective LLM cascading. Further, as indicated in section \ref{sec:appendix_prompts}, as the number and the complexity of tasks within the instruction increase, the model tend to have worse performance on the downstream task. As such, we propose to decompose the tasks within the instruction to several tasks and use different heads to handle it for achieving LLM cascade.

\section{Supplementary Results}
\subsection{Supplementary Cascade Results}
\label{sec:appendix_supplementary_cascade}
\begin{figure*}[ht]
    \centering
    \includegraphics[width=\textwidth]{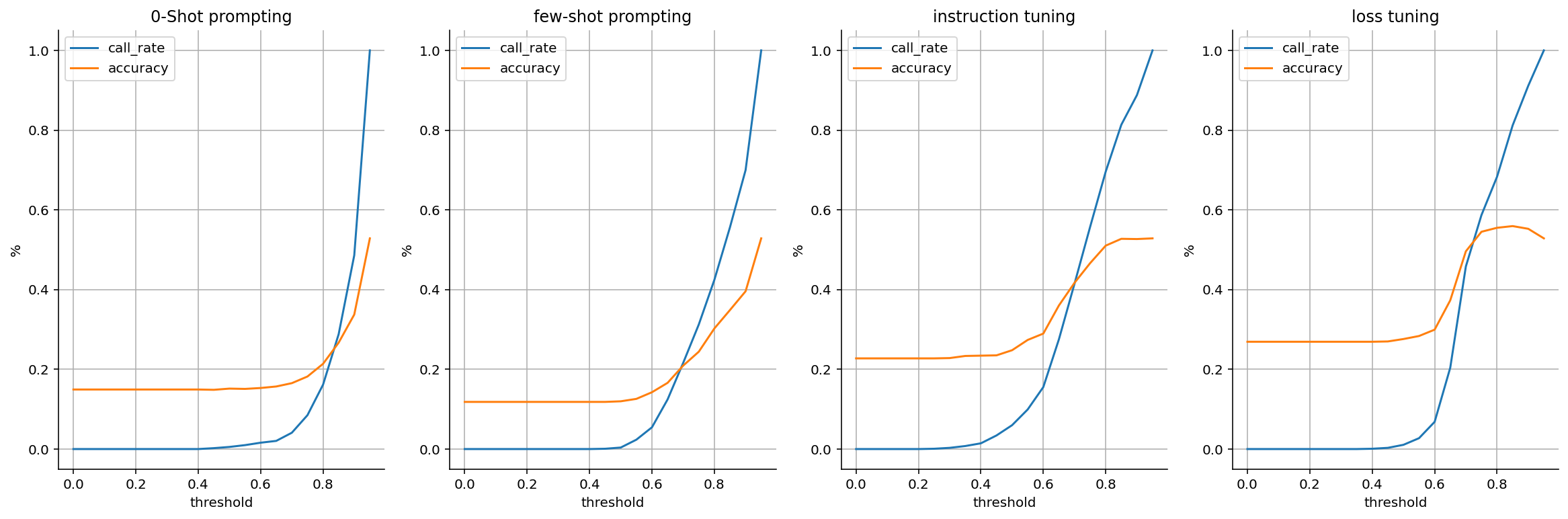}
    \caption{The curve of performance and call rate vs threshold on GSM8K dataset}
    \label{fig:th_cr_per}
\end{figure*}

\begin{figure*}[ht]
    \centering
    \subfigure[0-shot prompting]{\includegraphics[width=0.4\textwidth]{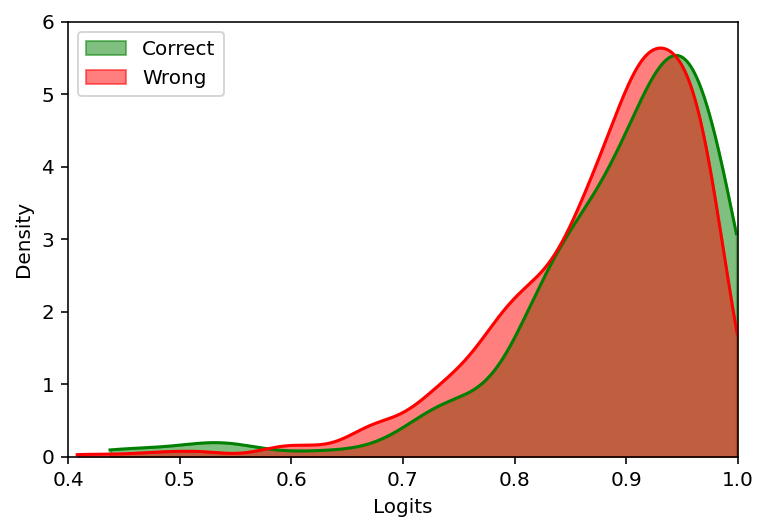}} 
    \subfigure[few-shot prompting]{\includegraphics[width=0.4\textwidth]{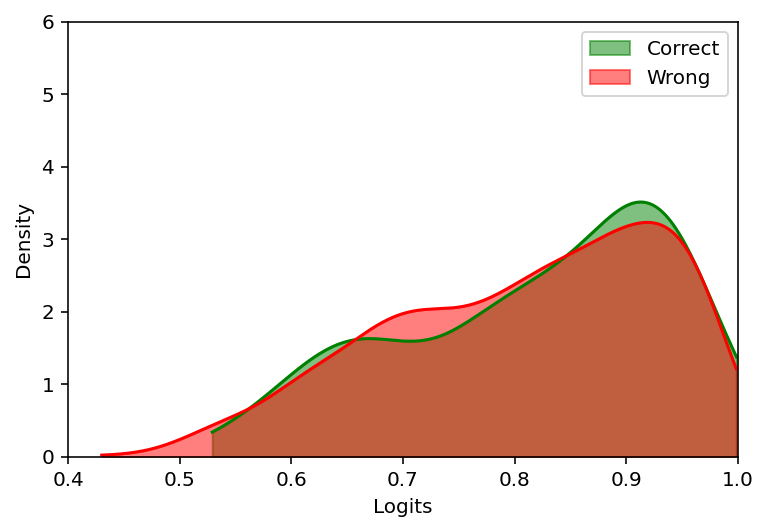}}
    \subfigure[instruction tuning]{\includegraphics[width=0.4\textwidth]{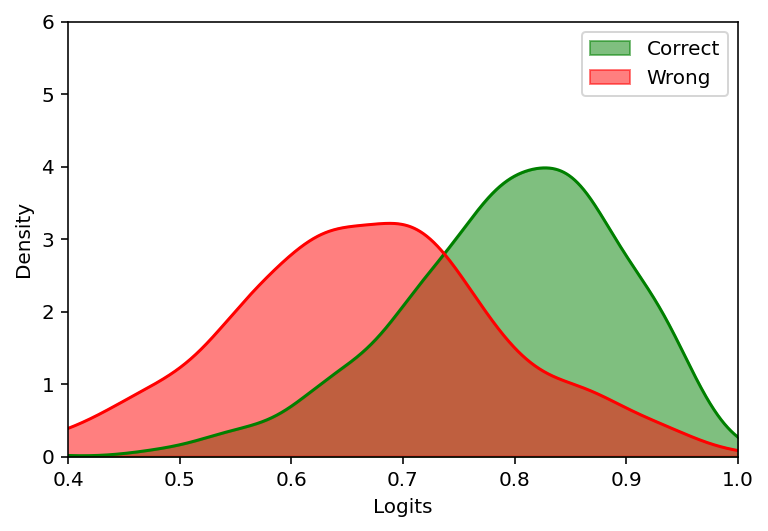}}
    \subfigure[loss tuning]{\includegraphics[width=0.4\textwidth]{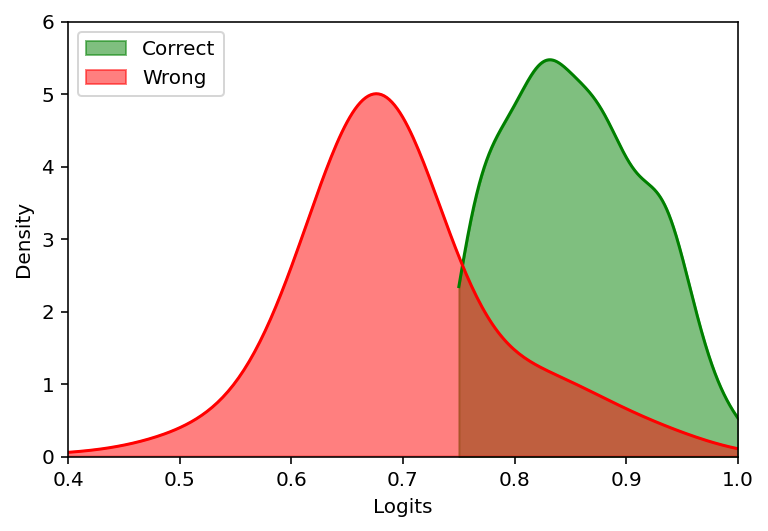}}
    \caption{Logits distribution curve by different methods on GSM8K dataset: (a) 0-shot prompting (b) few-shot prompting (c) instruction tuning (d) loss tuning.}
    \label{fig:logits_distribution}
\end{figure*}

As shown in Figure \ref{fig:logits_distribution}, training-based methods have a direct impact on distinguishing between correct and incorrect answers using logits (i.e., the separation between the green and red areas). This aligns with the scatter distribution in Figure \ref{fig:logits_distribution_scatter}, further validating the necessity of training in LLM cascading. Additionally, the higher peak in the red area indicates a faster performance improvement, as depicted in Figures \ref{fig:cascade_figure} and \ref{fig:th_cr_per}. These findings explain the effectiveness and intuition of our approach.

\subsection{Logits Distribution Study}
\begin{figure*}[ht]
    \centering
    \subfigure[0-shot prompting]{\includegraphics[width=0.3\textwidth]{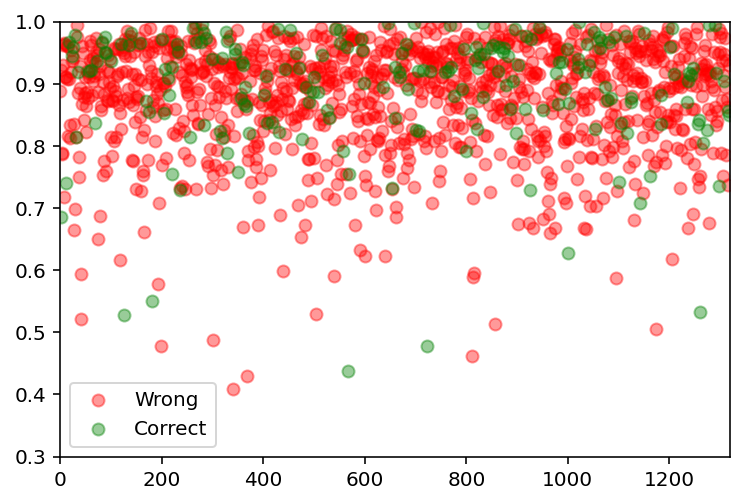}} 
    \subfigure[few-shot prompting]{\includegraphics[width=0.3\textwidth]{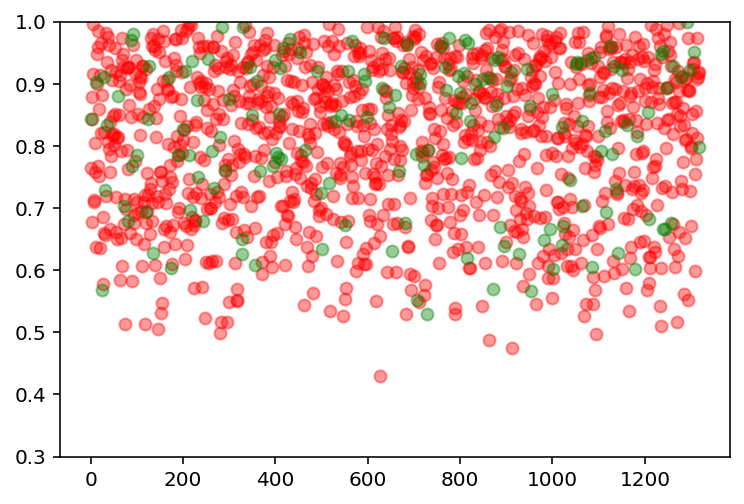}}
    \subfigure[few-shot prompting]{\includegraphics[width=0.3\textwidth]{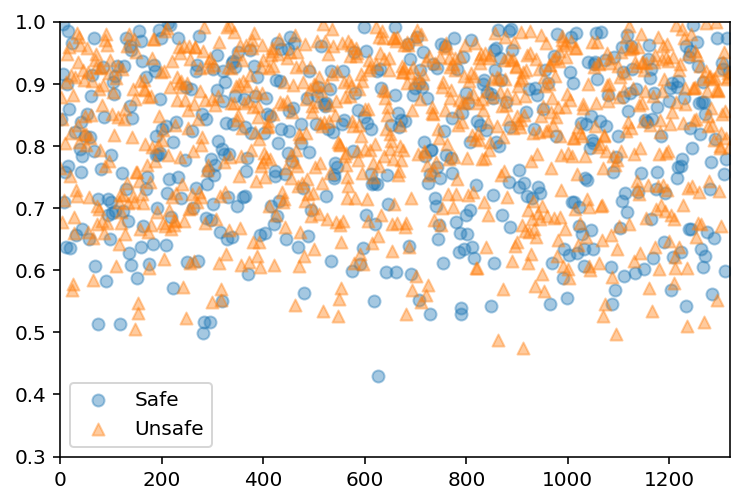}}
    \subfigure[instruction tuning]{\includegraphics[width=0.3\textwidth]{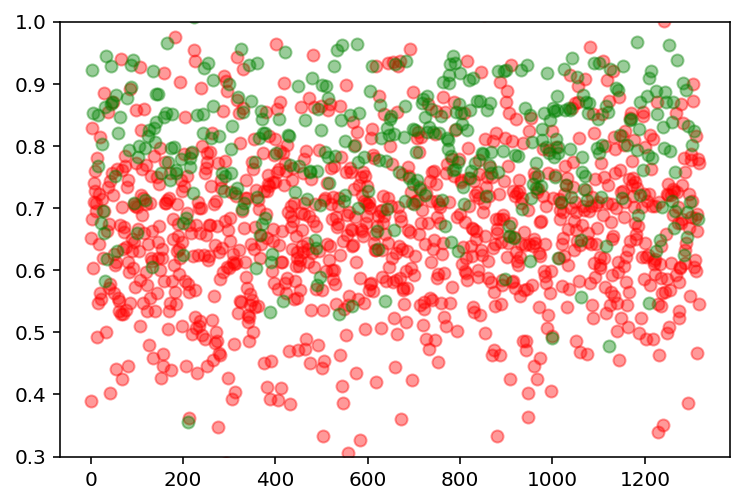}}
    \subfigure[loss tuning]{\includegraphics[width=0.3\textwidth]{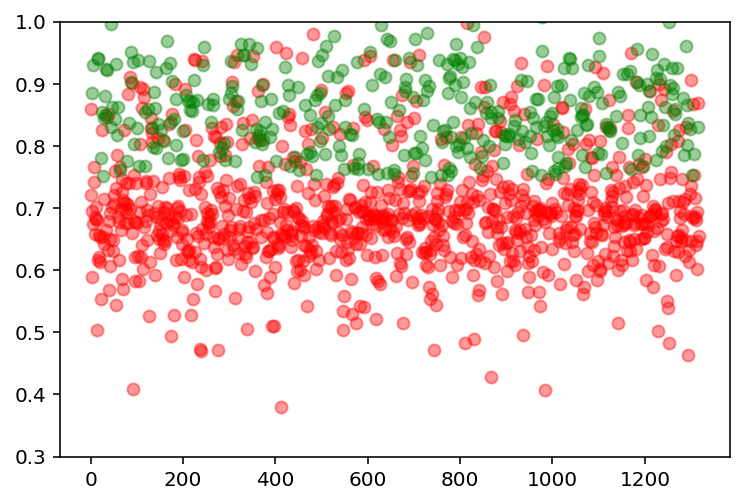}}
    \subfigure[loss tuning]{\includegraphics[width=0.3\textwidth]{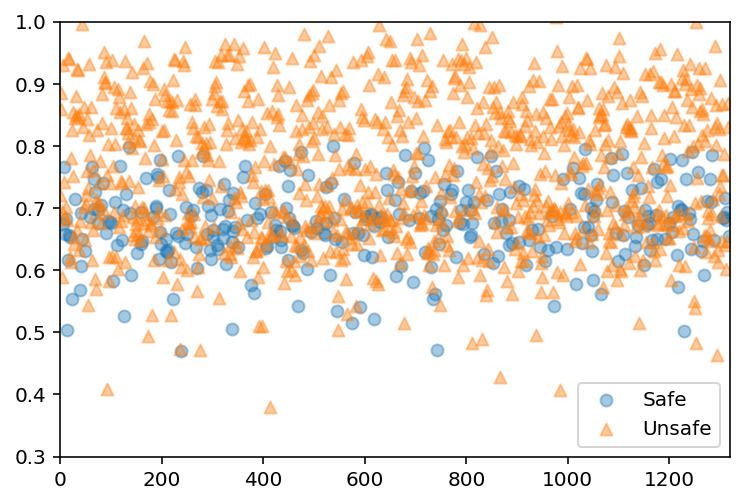}}
    \caption{Logits scatter distribution produced by different methods on GSM8K dataset. (e) and (f) are logits for privacy concerns; y-axis is the logits, x-axis is the data index.}
    \label{fig:logits_distribution_scatter}
\end{figure*}

To further understand the effectiveness of our proposed LLM cascade with multi-objective considerations, we visualize the logit distributions for both training and training-free methods. As shown in Figure \ref{fig:logits_distribution_scatter} and \ref{fig:logits_distribution}, the logits become more decentralized when a few examples are provided for $\Phi(L)$ to learn the cascade logic, in contrast to 0-shot prompting. Additionally, the signals within the distributions for prompting methods are not distinctly separable, which accounts for the randomness observed in routing queries, as discussed in previous sections. In contrast, training methods demonstrate more distinct distributions, where concentrated red points represent the reflection points noted in Figure \ref{fig:cascade_figure}. This indicates that training-based methods better grasp the cascade logic; answers with higher logits are correlated with more correct responses, suggesting that the trained $\Phi(L)$ is more confident in its correct answers and more likely to route difficult queries to the server. Furthermore, the trained model tends to send fewer unsafe queries to the server, as the logits for unsafe responses are generally higher, making them less likely to be sent. These observations reaffirm the effectiveness and necessity of incorporating multi-objective optimal considerations into cascading, highlighting the superiority of our proposed loss function for training the local LLM compared to existing prompting and instruction tuning methods.

\subsection{Datasets}
\label{sec:appendix_supplementary_datasets}
\begin{table*}[ht]
    \centering
    \begin{tabular}{cccc}
        \hline
        Dataset & GSM8K & MedSum & EmailSum \\
        \hline
        Avg. Input Length & 52.56 & 70.51 & 223.2 \\
        Avg. Output Length & 83.60 & 11.49 & 27.1 \\
        Avg. Leakage Tokens & 5.19 & 11.27 & 49.77 \\
        Task Type & Question Answering & Summarization & Summarization \\
        Measurement & Accuracy, Privacy Leakage & ROUGE, Privacy Leakage & ROUGE, Privacy Leakage \\
         \hline
    \end{tabular}
    \caption{Detailed type, statistics and measurement of datasets.}
    \label{tab:datasets}
\end{table*}
Table \ref{tab:datasets} provides detailed statistics for all datasets. Following the privacy research by \citet{hartmann2024can}, we extracted tokens with privacy concerns (e.g., names and other personal identifiers), as the number of such privacy-leakage tokens is critical for evaluating our methods. The extraction was based on PII rules \citep{kim2024propile} and HIPAA regulations \citep{lincke2024complying}, achieving extraction accuracies of 99.1\% for GSM8K and 99.7\% for MedQSum. A subset of 100 samples was manually verified by a highly educated PhD student, and the p-value score between human and machine extractions was less than 0.05, further validating the effectiveness of our proposed methods.

\section{Baseline Methodology}
\label{sec:appendix_methodology}
\begin{figure*}[h]
    \centering
    \includegraphics[width=\textwidth]{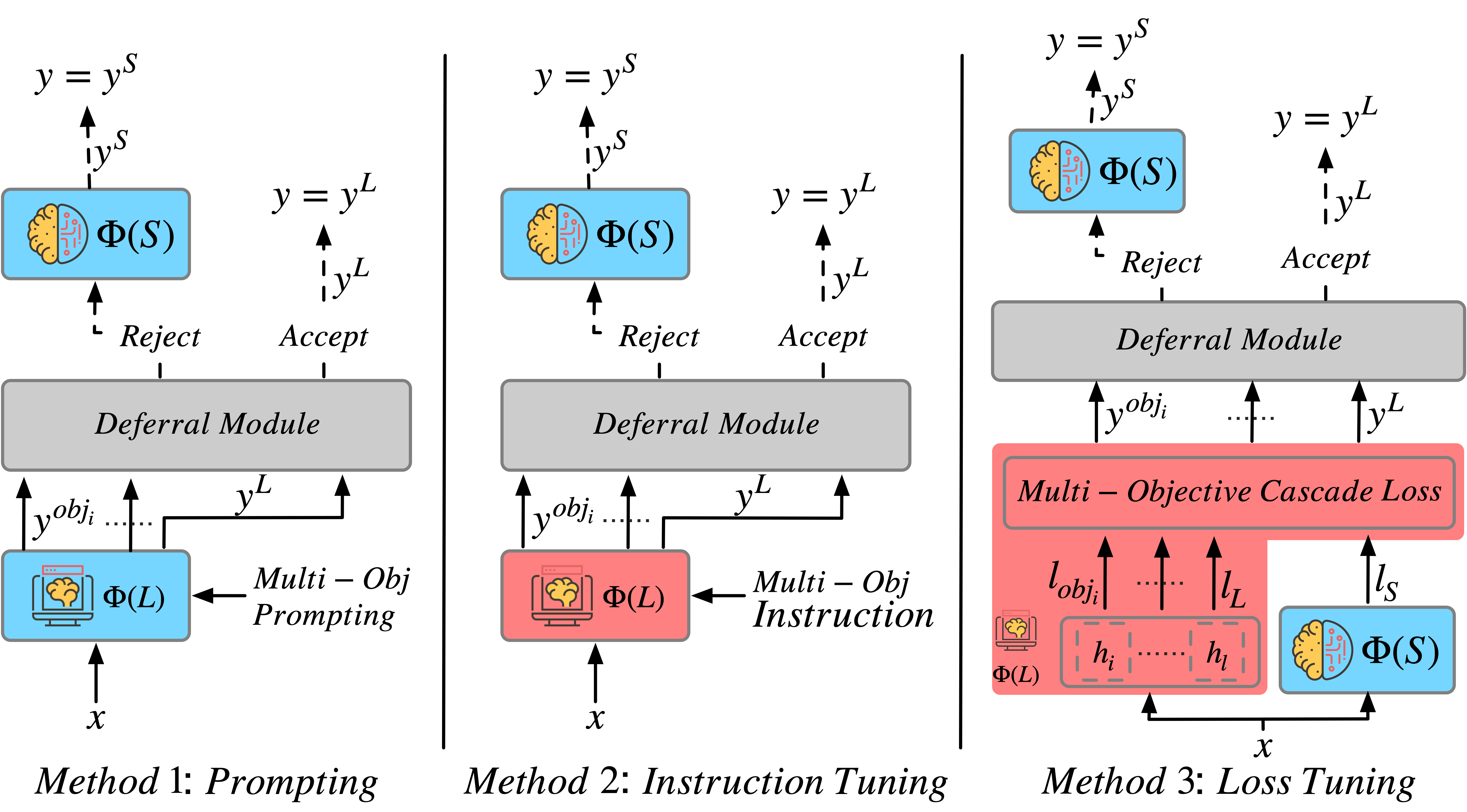}
    \caption{Overview of existing LLM cascade methods: $\Phi(L)$ and $\Phi(S)$ represent the local model and server model, respectively. The red box indicates trainable, while the blue box represents frozen. $\Phi(L)$ is tasked with generating responses $y^{L}$ and $y^{obj_{i}}$ for both the query $x$ and the multi-objective considerations $obj_{i}$. For loss tuning, the generation tasks are handled by different heads $h_{i}$, and a combined cascade loss is utilized for tuning.}
    \label{fig:methodology_apendix}
\end{figure*}
\subsection{Multi-Objective In-context Learning}
Ideally, the $\Phi(L)$ can be taught multi-objective optimal cascade logic based on its own natural language understanding ability. Efforts have been made to enable the $\Phi(L)$ being aware of the confidence of generated responses via self-critique\citep{zhu-etal-2024-towards}, step-by-step prompting\citep{zhang2023towards} etc. We step further on the previous works and include the privacy concern \citep{hartmann2024can} into prompt design. Specifically, we formulate an instructional prompt\footnote{The prompts used can be seen in the appendix \ref{sec:appendix_prompts}} which integrates query $x$ and objective considerations (i.e., privacy consideration $obj_{p}$) to the $\Phi(L)$ to obtain response $[y^{obj_{p}}, y^{L}]$, and these response will further be sent to the $D(\cdot)$ where deferral decisions will be made. Further, we follow \citet{deng2024wav2prompt}'s work and perform few-shot prompting to better activate the $\Phi(L)$'s in-context learning ability. However, with limited size, the $\Phi$ is inadequate\footnote{Please refer to the appendix \ref{sec:appendix_preliminary} for better understanding over the local llm's weakness.} to understand the multi-objective optimal cascade logic relying its own ability and the complicated logic may further hurt its ability to answer user's query and thus training is needed. 

\subsection{Multi-Objective Instruction Tuning}
Previous studies have demonstrated the effectiveness of instruction tuning in enhancing downstream task performance and improving comprehension of given instructions \citep{zhu-etal-2024-towards, zhao2024novel, ma2024llamoco, li2023reflection}. This ability to understand instructions aligns well with our objective of grasping the deferral logic. Furthermore, the improvements in task performance help mitigate any negative impacts on generating $y^{L}$ that may arise from producing $y^{obj_{i}}$ during prompting. Similar to the prompting method, we utilize an instructional prompt that combines a step-by-step instruction with the user query $x$ as input. The labeled text $\hat{y}$ corresponding to $x$, along with the labeled responses $\hat{y}^{obj_{i}}$ for the multi-objective considerations, serve as outputs for fine-tuning the model $\Phi(L)$. The responses generated by the tuned model will then be utilized by the deferral module $D(\cdot)$ to determine whether routing to the server model $\Phi(S)$ is necessary.

\subsection{Multi-Objective Loss Tuning}
Stepping further over the methods that rely on the local model's intricate understanding ability, recent works have pointed out the superiority of distilling the server llm's ability on downstream tasks into the loss function for tuning the local model\citep{wang2024cascade}. Intuitively, our assumption is that the server llm is larger and more powerful\citep{hartmann2024can} in terms of down-stream tasks, and thus the discrepancy between the generations of $\Phi(L)$ and $\Phi(S)$ can somehow be used for $\Phi(L)$ to indicate the confidence level. The larger the discrepancy is, the lower confidence level should the $\Phi(L)$ have. However, to enable $\Phi(L)$ being aware of multi-objective considerations, simply including the distillation loss from $\Phi(S)$ is inadequate. To this end, we decompose the overall task into several sub-tasks and use different heads to handle the different sub-tasks. Namely, given the multi-objective considerations $[obj_{1}, ..., obj_{i}]$ and the query $x$, we leverage multiple llm heads $[h_{1}, ..., h_{i}, h_{L}]$ to handle different considerations and the query. Each head will produce a loss and a distillation loss from $\Phi(S)$ will be optionally added. These losses will then be sent to a weighted-sum function to produce a multi-objective cascade loss for tuning $\Phi(L)$:
\begin{equation}
\begin{aligned}
    \centering
    &l = \sum\limits_{i}^{n} w_{i}\cdot l_{obj_{i}} + w_L \cdot l_{L} + \alpha(t) \cdot w_S \cdot l_{S} \\
    &\sum\limits_{i}^{n} w_{i}^{n} + w_{L} + w_{S}=1, \alpha(t) = H(logit_{y^{L}}, t)
\label{equation:multi-objCascsacdeLoss}
\end{aligned}
\end{equation}
where $w_{i}$ denotes the weight for the loss associated with generating response $y^{obj_{i}}$ for the objective $obj_{i}$, $w_{L}$ is the weight for the loss of generating response $y^{L}$ for $x$ from $\Phi(L)$ and $w_{S}$ is the weight for the loss of generating response $y^{S}$ for $x$ from $\Phi(S)$. $n$ is the number of objectives that need to be considered. $\alpha$ is the factor for controlling if the knowledge from the server LLM $\Phi(S)$ is used depending on a logit threshold $t$. $H(\cdot, t)$ is a modified Heaviside Step function which returns 0 if $\cdot>t$ else returns 1. In the context of identifying privacy concern, the loss function we utilized for tuning $\Phi(L)$ is:
\begin{equation}
\tiny
\begin{aligned}
    \centering
    l = &-w_{p}\cdot (\hat{y}^{p}\cdot log(p_{L}(y^{p}|x))+(1-\hat{y}^{p})\cdot log(1-p_{L}(y^{p}|x))) + \\
    &w_{L}\cdot log(p_{L}(y^{L}|x)) + \alpha(t)\cdot w_{S}\cdot log(p_{S}(y^{S}|x)) 
\label{equation:privacyCascsacdeLoss}
\end{aligned}
\end{equation}
where $y^{p}$, $\hat{y}^{p}$ are the predicted, golden binary predictions for privacy, respectively. Other terms remain the same as in formula \ref{equation:multi-objCascsacdeLoss}. By incorporating multi-objective considerations into the loss function for tuning $\Phi(L)$, the model will generate answers with better awareness of these considerations. The corresponding logits of the generated answers by tuned $\Phi(L)$ can then be utilized by the deferral module to inform decision-making. 
\subsection{Deferral Module}
All the three methods are studying how to enable the local LLM to be aware of multi-objective considerations while generating the response to the query. And such considerations are presented as the logit distributions of the generated response, for example, higher logit may indicated higher performance and less privacy concern. Deferral module plays a pivotal role in the LLM cascade since it decides which query to send out to the server llm based on the logits. Following previous successes on using different logit (e.g., mean, quantile) of the generated response as the reference to decide if there is a need to route the query to the server LLM\citep{wang2024cascade, jitkrittum2024does, gupta2024language}, we also utilize the logit of generated response as indicators to make the routing decisions. Specifically, given a threshold $t\in(0, 1)$, if the logit of the generated response exceed $t$ then it means the local LLM is confident with its response and no need to route, otherwise route the query $x$ to the server LLM $\Phi(S)$.

\end{document}